\documentclass[conference]{IEEEtran}
\IEEEoverridecommandlockouts
\usepackage{cite}
\usepackage{amsmath,amssymb,amsfonts}
\usepackage{algorithmic}
\usepackage{graphicx}
\usepackage{textcomp}
\usepackage{xcolor}
\def\BibTeX{{\rm B\kern-.05em{\sc i\kern-.025em b}\kern-.08em

    T\kern-.1667em\lower.7ex\hbox{E}\kern-.125emX}}
    
\usepackage{hyperref}
\hypersetup{breaklinks=true}

\usepackage{tipa}
\usepackage{multirow}
\definecolor{mycolor1}{HTML}{720056}
\definecolor{mycolor2}{HTML}{487984}
\definecolor{mycolor3}{HTML}{F5F0ED}
\usepackage{hyperref}
\hypersetup{
    pdfnewwindow=true,      
    colorlinks=true,        
    linkcolor=magenta,  
    citecolor=cyan,  
    filecolor=magenta,      
    urlcolor=purple    
}
\usepackage[T1]{fontenc}

\begin{document}

\title{ LLM-Powered Grapheme-to-Phoneme Conversion: Benchmark and Case Study } 

\author{

\IEEEauthorblockN{1\textsuperscript{st} Mahta Fetrat Qharabagh }
\IEEEauthorblockA{\textit{ Dept. of Computer Engineering } \\
\textit{Sharif University of Technology}\\
m.fetrat@sharif.edu }
\and
\IEEEauthorblockN{2\textsuperscript{nd} Zahra Dehghanian}
\IEEEauthorblockA{\textit{Dept. of Computer Engineering} \\
\textit{Sharif University of Technology}\\
zahra.dehghanian97@sharif.edu}
\and
\IEEEauthorblockN{3\textsuperscript{rd} Hamid R. Rabiee}
\IEEEauthorblockA{\textit{Dept. of Computer Engineering} \\
\textit{Sharif University of Technology}\\
rabiee@sharif.edu}

}

\maketitle

\begin{abstract}

Grapheme-to-phoneme (G2P) conversion is critical in speech processing, particularly for applications like speech synthesis. G2P systems must possess linguistic understanding and contextual awareness of languages with polyphone words and context-dependent phonemes. Large language models (LLMs) have recently demonstrated significant potential in various language tasks, suggesting that their phonetic knowledge could be leveraged for G2P. In this paper, we evaluate the performance of LLMs in G2P conversion and introduce prompting and post-processing methods that enhance LLM outputs without additional training or labeled data. We also present a benchmarking dataset designed to assess G2P performance on sentence-level phonetic challenges of the Persian language. Our results show that by applying the proposed methods, LLMs can outperform traditional G2P tools, even in an underrepresented language like Persian, highlighting the potential of developing LLM-aided G2P systems.

\end{abstract}

\begin{IEEEkeywords}
Grapheme-to-Phoneme Conversion (G2P), Large Language Models (LLMs), Prompt Engineering, Benchmarking Dataset, Context-Sensitive Phonemes
\end{IEEEkeywords}

\section{Introduction}

Grapheme-to-phoneme (G2P) conversion is a well-established task in speech processing and linguistics, serving as a critical component in modern Text-to-Speech (TTS) models like FastSpeech2 \cite{ren2020fastspeech}. However, G2P conversion becomes significantly more challenging in languages with polyphone words and context-sensitive phonemes. For example, the English word "read" can be pronounced as /\textit{ri:d}/ or /\textit{red}/ depending on the context. Similarly, This issue arises in Persian and Arabic because short vowels, known as diacritics, are often omitted in written form, making pronunciation ambiguous. For example, the Persian words for "flower" and "mud" are written the same but pronounced differently as /\textit{gol}/ and /\textit{gel}/. Another context-sensitive feature in Persian is the Ezafe, 
a short vowel /\textit{e}/ primarily used to connect two or more related words in a noun phrase, showing possession, description, or relationship between elements. For example, the sentences "This is Ziba's flower" and "This flower is beautiful" have identical written forms but differ in pronunciation due to the Ezafe:

\begin{enumerate}
    \item /\textit{in gol-e zibA ast}/
    \item /\textit{in gol zibA ast}/
\end{enumerate}

As illustrated, the presence or absence of the Ezafe phoneme /\textit{e}/, determined by context, can change the sentence's meaning completely. This context dependency is so subtle that even native speakers may struggle to identify it without additional clues from surrounding sentences. These complexities underscore the need for G2P tools incorporating linguistic knowledge and contextual awareness.

Large language models (LLMs) have recently demonstrated significant linguistic and phonetic capabilities \cite{suvarna2024phonologybench}, making them promising candidates for G2P tasks. However, several challenges arise when applying LLMs to this task. First, their outputs can still contain substantial errors. Second, LLMs often perform poorly in underrepresented languages, as their effectiveness depends on the availability of phonetic data and resources in training data of those languages.
\begin{table*}[t]
\centering
\caption{\textbf{Performance of available Persian G2P models.} "N/D" indicates that the model made no positive predictions, making precision undefined.
}
\label{tab:old_g2p_performance}
\scalebox{1.05}{
\begin{tabular}{lccccccc}
\hline

\hline 
 & \multicolumn{1}{c}{\cite{persian-phonemizer}} & \multicolumn{1}{c}{\cite{PasaOpasen}} & \multicolumn{1}{c}{\cite{azamrabiee}} & \multicolumn{1}{c}{\cite{mohammadhasan}} & \multicolumn{1}{c}{\cite{Mortensen-et-al:2018}} & \multicolumn{1}{c}{\cite{sajadalipour7}}  & \multicolumn{1}{c}{\textbf{}}\\
      & \multicolumn{1}{c}{\textbf{persian-phonemizer}} & \multicolumn{1}{c}{\textbf{PersianG2P}} & \multicolumn{1}{c}{\textbf{Persian\_G2P}} & \multicolumn{1}{c}{\textbf{G2P}} & \multicolumn{1}{c}{\textbf{Epitran}} & \multicolumn{1}{c}{\textbf{G2P with Transformer}}
 & \multicolumn{1}{c}{\textbf{Ours}}\\
\hline

\textbf{PER (\%) ↓}  & $27.89$ & $17.91$ & $35.83$ & $20.13$ & $47.53$ & \underline{$15.37$} & $\mathbf{8.30}$\\
\textbf{Polyphone Acc. (\%) ↑} & $29.00$ & $37.50$ & $20.50$ & $30.00$ & $0.00$ & \underline{$40.50$} & $\mathbf{54.00}$\\
\textbf{Ezafe F1 (\%) ↑} & \underline{$62.06$} & $14.69$ & $6.01$ & $17.31$ & N/D & $11.98$ & $\mathbf{88.33}$\\
\hline
\end{tabular}
}
\end{table*}

In this work, we explore various prompting techniques and introduce post-processing methods to maximize the performance of existing LLMs without additional training. We also introduce a benchmarking dataset to evaluate G2P tools on sentence-level phonetic challenges in Persian. Our results indicate that LLM-based G2P has significant potential, enabling the creation of stand-alone G2P tools by generating large-scale, sentence-level annotated datasets by LLMs without human intervention. Key contributions of our work include:

\begin{itemize}
\item Presenting the first LLM-based G2P tool.
\item Proposing innovative prompting and post-processing techniques to improve LLMs' phonetic accuracy in G2P tasks.
\item Creating the first sentence-level benchmarking dataset, "Sentence-Bench", and developing an Ezafe prediction metric to evaluate G2P models on sentence-level challenges in Persian.
\item Introducing "Kaamel-Dict," the largest open-licensed G2P Persian dictionary. 
\item Benchmarking state-of-the-art LLMs in sentence-level G2P task using the proposed methods.
\end{itemize}

\section{Related Works}

Grapheme-to-phoneme (G2P) conversion has a long history, beginning with instance-based learning \cite{van1993data}, rule-based approaches \cite{kim2002morpheme}, Hidden Markov Models \cite{taylor2005hidden}, and joint-sequence models \cite{bisani2008joint}, and progressing to more recent deep neural network models such as LSTMs \cite{rao2015grapheme}, CNNs \cite{yolchuyeva2019grapheme}, CTC \cite{wang2023liteg2p}, and Transformers \cite{sevinj2019transformer, yu2020multilingual, vrezavckova2024t5g2p}. However, none of these methods have leveraged the valuable phonological and linguistic knowledge embedded in large language models (LLMs) in their approaches.

There have been a few studies that touch on the phonological capabilities of LLMs. One work evaluated LLMs on a simple phonological task among other linguistic challenges \cite{beguvs2023large}, another assessed LLMs’ phonological abilities using a multiple-choice question dataset \cite{peng2023spoken}, and only one study \cite{suvarna2024phonologybench} directly evaluated LLMs on the G2P task, alongside two other tasks, concluding that LLMs were inferior to traditional models in this domain.

However, this last study had two significant limitations in fully capturing the potential of LLMs for G2P. First, it evaluated G2P at the word level, overlooking the key advantage of LLMs in sentence-level processing and context awareness. Second, the prompting method used was overly simplistic. We show that directly prompting LLMs for phonetic transcriptions could be counterproductive, and that even a simple mapping on the outputs could substantially improve results.

To the best of our knowledge, no scholarly work has been published to date in Persian G2P literature. Although some models exist \cite{persian-phonemizer, sajadalipour7, azamrabiee, mohammadhasan, PasaOpasen}, they are trained solely at the word level, despite the context-sensitive nature of Persian phonetics, and rely on limited dictionaries to address out-of-vocabulary cases.

\section{Datasets and Benchmarks}

\subsection{G2P Dictionary}\label{DICT}
The existing G2P tools for Persian are trained on several dictionaries, primarily sourced from Tihu-Dict \cite{tihu-dict}, IPA-Dict \cite{ipa-dict}, and Wiktionary \cite{wiktionary}. The size of these dictionaries varies, ranging from approximately 2,000 \cite{azamrabiee} to 54,000 \cite{zaya} grapheme-phoneme pairs. Additionally, we identified a new resource from the Persian Jame Glossary \cite{jame}. However, these datasets use different phonetic representation formats, making it difficult to leverage them collectively to develop a robust G2P system.

To address this, we developed a module that unifies the phonetic representations across these datasets and merged all the dictionaries from the existing G2P resources \cite{mohammadhasan, PasaOpasen, zhu2022charsiu-g2p, IPA-Translator} with the newly collected one. This effort resulted in the "Kaamel-Dict", a dataset containing over 120,000 entries, making it the largest Persian G2P dictionary to date. We are releasing this dataset under an open GNU license, as all the source dictionaries are licensed for redistribution.\footnote{https://huggingface.co/datasets/MahtaFetrat/KaamelDict}

\subsection{Sentence-Level G2P Benchmarking Dataset}

To the best of our knowledge, no dataset exists with phoneme-annotated sentences in Persian for evaluating G2P models at the sentence level. Furthermore, two specific challenges in sentence-level phoneme translation require more focused evaluation: 1) predicting the correct pronunciation of polyphone words within a sentence, and 2) predicting context-sensitive phonemes such as the Ezafe.

To address these challenges, we created the first sentence-level benchmarking dataset, named "Sentence-Bench," in two parts. First, we selected 100 of the most upvoted sentences from CommonVoice \cite{ardila2019common} and 100 random sentences from the ManaTTS \cite{manatts} dataset. These sentences were annotated with their phonetic sequences.

For the second part, we selected approximately 100 polyphone words from the "Kaamel" dictionary and manually constructed 200 sentences containing these words in various pronunciations. Each sentence in this part includes the phoneme sequence for the entire sentence, the polyphone word, and its correct pronunciation within the sentence. The resulting dataset, comprising 400 sentences, is released under a GNU license as permitted by the dictionary license.\footnote{https://huggingface.co/datasets/MahtaFetrat/SentenceBench}

Additionally, to evaluate G2P models on Ezafe prediction, we propose a new metric and have developed a module to detect Ezafe phonemes in the model output and assess the model’s performance on Ezafe prediction. This dataset enables the evaluation of G2P tools on sentence-level challenges using three metrics: 1) the conventional PER, 2) accuracy in predicting the correct pronunciation of polyphone words, and 3) performance in Ezafe prediction.

\section{Methods}
\label{sec:methods}

\begin{table*}[htbp]
\centering
\caption{\textbf{Performance of llama-3.1-405b-instruct using the different methods of section \ref{sec:methods}. }
}
\label{tab:methods_performance}
\scalebox{1.03}{
\begin{tabular}{lcccccccccc}
\hline
 & \textbf{1) Naive} & \textbf{2) ICL} & \textbf{3) Finglish} & \textbf{4) Rule-based} & \textbf{5) LLM-based} & \multicolumn{3}{c}{\textbf{6) Dict Hints}} & \textbf{7) Combined} \\
\cline{7-9}
 & & & & \textbf{Correction} & \textbf{Correction} & \textbf{(1)} & \textbf{(2)} & \textbf{(3)} & \\
\hline
\textbf{PER (\%) ↓}            & $31.61$ & $15.79$ & $10.86$ & $10.25$ & $8.44$ & $9.88$ & \underline{$8.30$} & $\mathbf{8.26}$ & $8.59$ \\
\textbf{Polyphone Acc. (\%) ↑} & $23.50$ & $37.00$ & $56.00$ & $56.50$ & \underline{$58.00$} & $56.50$ & $54.00$ & $54.00$ & $\mathbf{61.00}$ \\
\textbf{Ezafe Accuracy (\%) ↑} & $88.33$ & $95.90$ & \underline{$96.54$} & $95.62$ & $\mathbf{96.85}$ & \underline{$96.54$} & $\mathbf{96.85}$ & $95.90$ & $96.11$ \\
\textbf{Ezafe Precision (\%) ↑}& $72.41$ & $89.34$ & $\mathbf{91.07}$ & $84.13$ & \underline{$90.31$} & $87.70$ & $89.56$ & $85.23$ & $86.27$ \\
\textbf{Ezafe Recall (\%) ↑}   & $23.70$ & $79.46$ & $82.84$ & $83.75$ & $86.23$ & \underline{$86.91$} & $\mathbf{87.13}$ & $84.65$ & $85.10$ \\
\textbf{Ezafe F1 (\%) ↑} & $35.71$ & $84.11$ & $86.76$ & $83.94$ & \underline{$88.22$} & $87.30$ & $\mathbf{88.33}$ & $84.94$ & $85.68$ \\
\hline
\end{tabular}
}
\end{table*}

\begin{table*}[t]
\centering
\caption{\textbf{Performance of different LLMs on the test dataset using the method selected in section \ref{sec:methods}. }
}
\label{tab:llms_performance}

\scalebox{1.01}{
\begin{tabular}{lccccccccc}

\hline 
 & \multicolumn{1}{c}{\textbf{llama-3.1}} & \multicolumn{1}{c}{\textbf{gemma2}} & \multicolumn{1}{c}{\textbf{mixtral}} & \multicolumn{1}{c}{\textbf{qwen-2}} & \multicolumn{1}{c}{\textbf{mistral}} & \multicolumn{1}{c}{\textbf{gpt-3.5}} & \multicolumn{1}{c}{\textbf{gpt-4o}} & \multicolumn{1}{c}{\textbf{gpt-4}} & \multicolumn{1}{c}{\textbf{claude-3.5}} \\
 & \multicolumn{1}{c}{\textbf{405b-instruct}} & \multicolumn{1}{c}{\textbf{9b-it}} & \multicolumn{1}{c}{\textbf{8x7b}} & \multicolumn{1}{c}{\textbf{7b-instruct}} & \multicolumn{1}{c}{\textbf{7b-instruct}} & \multicolumn{1}{c}{\textbf{turbo-instruct}} & \multicolumn{1}{c}{\textbf{mini}} & \multicolumn{1}{c}{} & \multicolumn{1}{c}{\textbf{sonnet}} \\
\hline

\textbf{PER (\%) ↓}            & $8.30$ & $21.58$ & $26.84$ & $59.06$ & $34.68$ & $11.76$ & $10.44$ & \underline{$8.28$} & $\mathbf{5.80}$ \\
\textbf{Polyphone Acc. (\%) ↑} & \underline{$54.00$} & $21.50$ & $15.00$ & $3.50$ & $12.50$ & $40.50$ & $45.00$ & $48.50$ & $\mathbf{78.50}$ \\
\textbf{Ezafe F1 (\%) ↑}   & \underline{$88.33$} & $61.21$ & $44.05$ & $27.99$ & $38.93$ & $73.04$ & $70.34$ & $87.26$ & $\mathbf{93.03}$ \\
\hline
\end{tabular}
}
\end{table*}

While LLMs have demonstrated an understanding of the phonemic representations of languages, the effectiveness of their output heavily depends on the quality of the prompting techniques employed. These techniques involve the formulation of queries and the inclusion of relevant information and context. 

This study explored various prompting strategies to optimize LLM performance in G2P tasks. This section outlines our methodical approach, beginning with the most basic strategy and progressively integrating more sophisticated techniques. We report our findings using Meta's recent LLaMA 405b model \cite{meta2024llama3}.

\begin{enumerate} \item \textbf{Naive Phonetic Approach:} In our initial method, we prompted the model to generate the International Phonetic Alphabet (IPA) representation of a given Persian sentence (Fig.~\ref{fig:methods1-3}).
We then mapped the model’s output to a simplified phonetic notation commonly used by existing G2P tools and dictionaries for Persian. This mapping step is crucial because LLMs often produce varied and unconventional symbols when generating phonetic transcriptions. For instance, a model might use any of the symbols \textit{a, ā, ä, â, á, æ} for the Persian short vowel /æ/. This naive approach resulted in an average Phoneme Error Rate (PER) of $31.61\%$ on our test dataset.

\begin{figure}[htbp]
\centering
\includegraphics[width=0.65\columnwidth]{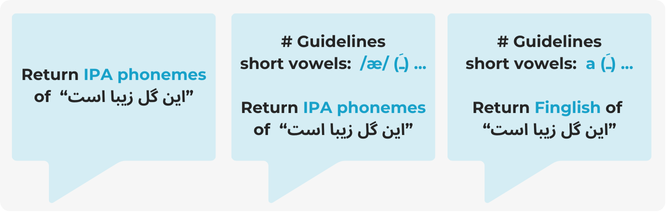}
\caption{Schematic of prompts for Methods 1 (left), 2 (middle), and 3 (right).}
\label{fig:methods1-3}
\end{figure}

\item \textbf{In-context Learning:} As a first improvement, we explored one-shot, few-shot, and in-context learning techniques, where the model was provided with examples and more detailed explanations of the output representations and linguistic rules. The model achieved PERs of $19.89\%$, $17.19\%$, and $15.79\%$ for these methods, respectively. Based on these results, we opted to continue with in-context learning in subsequent experiments (Fig.~\ref{fig:methods1-3}).

\item \textbf{Finglish:} Persian speakers sometimes type Persian (Farsi) words using the Latin alphabet, a practice known as Finglish. Unlike Persian script, where short vowels (diacritics) are typically omitted, Finglish often includes phonemes representing these vowels, making it a phonetic approximation of Persian sentences. Since LLMs likely have more exposure to Finglish data than to IPA phonetic representation, they may be better at generating Finglish text than producing standard phonetic forms. We modified our prompts to request the Finglish form and then mapped the output back to the desired phonetic representation (Fig.~\ref{fig:methods1-3}). However, this mapping is challenging due to the inherent ambiguity of Finglish. For instance, the Persian phoneme /\ \textesh\ / is represented as "sh" in Finglish, which could also be interpreted as the two phonemes /\textit{s}/ and /\textit{h}/. The mapping module
leverages linguistic rules to address such challenges. This approach achieved a PER of $10.86\%$ on the converted phonetic representation, leading us to use Finglish as the preferred output format in subsequent methods.

\item \textbf{Rule-based Dictionary Correction: } In this step, we aimed to correct minor spelling errors in the LLM output. First, we searched the grapheme words in the dictionary to retrieve their exact phonetic representations. We then employed word similarity metrics to match the phonetic representation predicted by the LLM with those in the dictionary (Fig.~\ref{fig:method4}). This process reduced the model’s PER to $10.25\%$.

\begin{figure}[htbp]
\centering
\includegraphics[width=0.65\columnwidth]{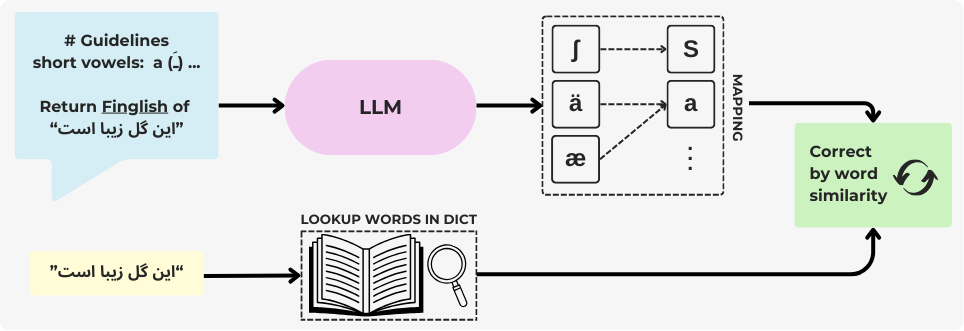}
\caption{Schematic of Method 4.}
\label{fig:method4}
\end{figure}
    
\item \textbf{LLM-based Dictionary Correction:} Despite improvements, the rule-based method using word similarity metrics is not always reliable. To enhance accuracy, we implemented a method to correct the LLM's phonetic output by re-prompting the model, including potential phonetic representations of words in the prompt (Fig.~\ref{fig:method5}). This approach further reduced the PER to $8.44\%$.

\begin{figure}[htbp]
\centering
\includegraphics[width=0.65\columnwidth]{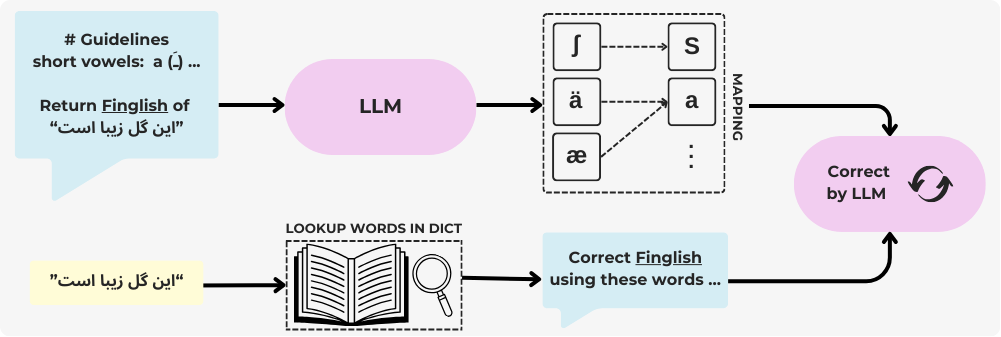}
\caption{Schematic of Method 5.}
\label{fig:method5}
\end{figure}

\item \textbf{Dictionary Hints: } In this method, instead of relying on post-processing to correct the model’s output using the dictionary, we embedded dictionary information directly into the initial prompt. This was implemented in three ways: (1) by suggesting all possible phonetic alternatives for each word, (2) by providing the phonetics for unambiguous, single-phone words as hints, and (3) by substituting the phonetic representation of these words directly within the grapheme and prompting the model to complete the remaining phonetics (Fig.~\ref{fig:methods-6}). The respective PERs for these approaches were $9.88\%$, $8.30\%$, and $8.26\%$. Although the PERs are relatively close, the methods vary in performance across other metrics, as detailed in Table~\ref{tab:methods_performance}.

\begin{figure}[htbp]
\centering
\includegraphics[width=0.75\columnwidth]{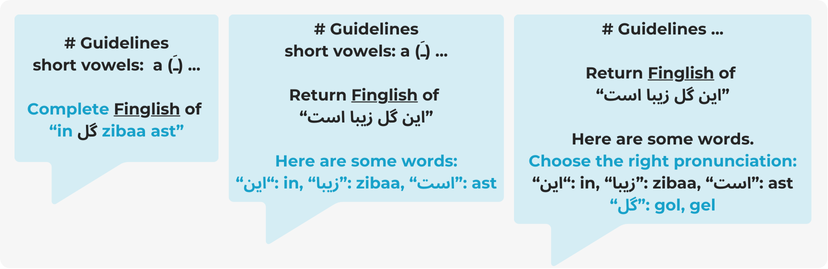}
\caption{Schematic of prompts for Methods 6 (1) (left), (2) (middle), and (3) (right).}
\label{fig:methods-6}
\end{figure}

\item \textbf{Combined Methods: } We explored various combinations of the previously discussed techniques to create composite methods, aiming for even better results. The most effective strategy combined elements from the Dictionary Hints and LLM-based Correction methods. Specifically, the model was first prompted to complete the phonetic representation of a sentence, followed by a second prompt that asked it to refine the output based on the dictionary-provided phonetics for specific words. This approach significantly improved the model’s performance with polyphone words, as shown in Table~\ref{tab:methods_performance}.

\end{enumerate}

\begin{figure}[htbp]
\centering
\includegraphics[width=0.9\columnwidth]{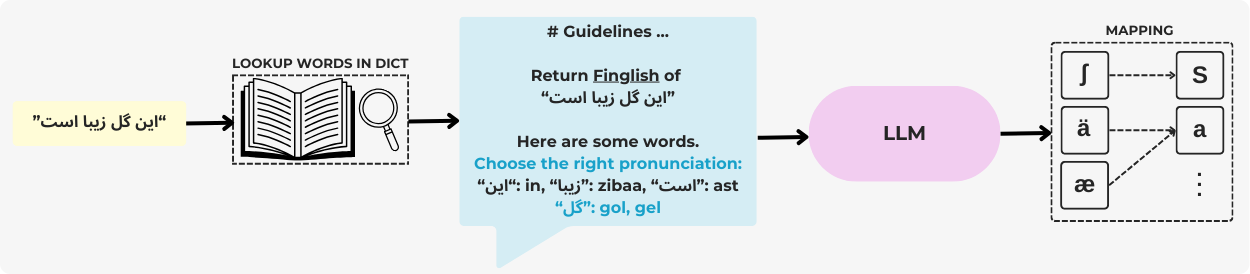}
\caption{Schematic of the final method.}
\label{fig:method6-2}
\end{figure}

The Ezafe phoneme is present in most Persian sentences, making its accurate prediction a key indicator of a model's contextual awareness. Consequently, we selected the final method based on its performance in Ezafe prediction. As detailed in the next section, method 6-(2) (Fig.~\ref{fig:method6-2}) delivers the best results on this metric and ranks second in terms of PER. This method is also preferred over methods 5 and 7 due to its use of a single prompt, reducing both cost and latency. In the following section, we benchmark various LLMs using this approach.

\section{Results}	

We first evaluated the current Persian G2P models. Table~\ref{tab:old_g2p_performance}
 summarizes their performance in terms of Phoneme Error Rate (PER), Ezafe prediction F1 score, and exact polyphone word prediction on the "Sentence-Bench" dataset. The "Ours" column presents the best results from our G2P, using the top open-source LLM, LLama 3.1 405b. In all the tables, the best results are in bold, and the second best results are underlined.

Next, we evaluated various methods using a single LLM, LLama 3.1 405b. The results are shown in Table~\ref{tab:methods_performance}.

Finally, we evaluated the selected method across various state-of-the-art LLMs, including GPT-3.5 \cite{openai2022chatgpt}, GPT-4 \cite{openai2023gpt4}, and Claude 3.5 Sonnet-20240620 \cite{claude3-anthropic}, as well as multiple open-source LLMs such as LLaMA 405 \cite{meta2024llama3}, Gemma2 9b \cite{team2024gemma}, Mixtral 8x7b-32768 \cite{jiang2024mixtral}, qwen2 7b \cite{bai2023qwen}, and mistral 7b \cite{jiang2023mistral}. The results are summarized in Table~\ref{tab:llms_performance}.

\section{Discussion}

Our experiments reveal that while the raw outputs of LLMs may initially fall short compared to traditional G2P models, the proposed prompting techniques significantly enhance their performance. When utilizing our method, the LLMs not only surpassed traditional models in terms of Phoneme Error Rate (PER), but they also demonstrated superior accuracy in handling polyphone words. This suggests that LLMs possess a deeper contextual understanding, which is crucial for languages like Persian where phonetic ambiguities are common. Additionally, the results for Ezafe detection are highly promising, helping to address the issue of missing Ezafe phonemes in Persian TTS, which often leads to robotic-sounding speech.

The strong performance of LLM-based tools in sentence-level G2P tasks, as presented in this study, underscores two key implications. First, LLMs have the potential to serve as the foundation for highly effective G2P models in online settings. Second, they can be invaluable in generating large-scale datasets of sentences with automated phonetic labels, which can then be used to train offline G2P models. This capability could significantly reduce the reliance on labor-intensive human annotation processes and enable the creation of much larger datasets. The offline models, imbued with the distilled knowledge of LLMs in G2P tasks, could then be integrated into modern TTS systems.

\section{Conclusion}
In this work, we introduced a framework that significantly enhances LLM performance in G2P tasks, surpassing traditional models by a significant margin, and benchmarked state-of-the-art LLMs using this methodology. We also contributed the largest Persian G2P dictionary, "Kaamel-Dict," along with a benchmarking dataset, "Sentence-Bench," and two additional metrics specifically designed for evaluating G2P models in sentence-level contexts. By releasing these resources under an open license, we aim to facilitate future research on the challenges posed by context-sensitive languages, particularly at the sentence level.

\bibliographystyle{IEEEbib}
\bibliography{refs}

\end{document}